\documentclass[runningheads]{llncs}
\usepackage{graphicx}
\usepackage{comment}
\usepackage{amsmath,amssymb} 
\usepackage{color}
\usepackage{epsfig}
\usepackage{times}  
\usepackage{helvet}  
\usepackage{courier}  
\usepackage{url}  
\usepackage{multirow}
\usepackage{tabularx} 
\usepackage{caption}
\usepackage{xspace}
\usepackage{booktabs}
\usepackage{subfigure}
\usepackage{enumitem}
\usepackage{paralist} 
\usepackage{epstopdf}
\usepackage{amsmath}

\def \ie {{i.e.}\xspace}
\def \eg {{e.g.}\xspace}

\def \etal {{et al.}\xspace}
\def \vs {{v.s.}\xspace}

\newcommand{\pms}[1]{\tiny{$\pm$ #1}}

\newcommand{\para}[1]{\noindent \textbf{#1}}

\begin{document}
\pagestyle{headings}
\mainmatter
\def\ECCVSubNumber{3548}  

\title{Active Crowd Counting with Limited Supervision}

%
\author{Zhen Zhao$^{1}$\thanks{Authors contributed equally.} \and Miaojing Shi$^{2\star}$ \and Xiaoxiao Zhao$^1$ \and Li Li$^{1,3}$}
%
\authorrunning{Zhao et al.}
%
\institute{$^1$ College of Electronic and Information Engineering, Tongji University\\
$^2$ King's College London\\
$^3$ Institute of Intelligent Science and Technology, Tongji University\\
\parbox{15cm}{\email{zhenzhao0917@gmail.com};~
\email{miaojing.shi@kcl.ac.uk};~
\email{lili@tongji.edu.cn}}}
\maketitle

\begin{abstract}
To learn a reliable people counter from crowd images, head center annotations are normally required. Annotating head centers is however a laborious and tedious process in dense crowds. In this paper, we present an active learning framework which enables accurate crowd counting with limited supervision: given a small labeling budget, instead of randomly selecting images to annotate, we first introduce an active labeling strategy to annotate the most informative images in the dataset and learn the counting model upon them. The process is repeated such that in every cycle we select the samples that are diverse in crowd density and dissimilar to previous selections. In the last cycle when the labeling budget is met, the large amount of unlabeled data are also utilized:
a distribution classifier is introduced to align the labeled data with unlabeled data; furthermore, we propose to mix up the distribution labels and latent representations of data in the network to particularly improve the distribution alignment in-between training samples.
We follow the popular density estimation pipeline for crowd counting. Extensive experiments are conducted on standard benchmarks \ie ShanghaiTech, UCF\_CC\_50, MAll, TRANCOS, and DCC. By annotating limited number of images (\eg 10\% of the dataset), our method reaches levels of performance not far from the state of the art which utilize full annotations of the dataset.
\end{abstract}

\section{Introduction}\label{Sec:introducation}
The task of crowd counting in computer vision is to
automatically count people numbers in images/videos. With the rapid growth of world's population, crowd gathering becomes more frequent than ever. To help with crowd control and public safety, accurate crowd counting is demanded.

Early methods count crowds via the detection of individuals~\cite{viola2003ijcv,brostow2006cvpr,rabaud2006cvpr}. They suffer from heavy occlusions in dense crowds. More importantly, learning such people detectors normally requires bounding box or instance mask annotations for individuals, which often makes it undesirable in large-scale applications. Modern methods mainly conduct crowd counting via density estimation~\cite{onoro2016eccv,zhang2016cvpr,sindagi2017iccv,sam2018aaai,zhang2018wacv,liu2018cvpr,li2018cvpr,xu2019iccv}. Counting is realized by estimating a density map of an image whose integral over the image gives the total people count.
Given a training image, its density map is obtained via Gaussian blurring at every head center. Head centers are the required annotations for training. Thanks to the powerful deep neural networks (DNNs)~\cite{krizhevsky2012nips}, density estimation based methods show a great success in recent progress~\cite{zhang2016cvpr,sam2017arxiv,li2018cvpr,ranjan2018eccv,shi2019cvpr,xu2019iccv,shi2019iccv}.

\begin{figure}[t]
	\centering
	\includegraphics[width=0.6\columnwidth]{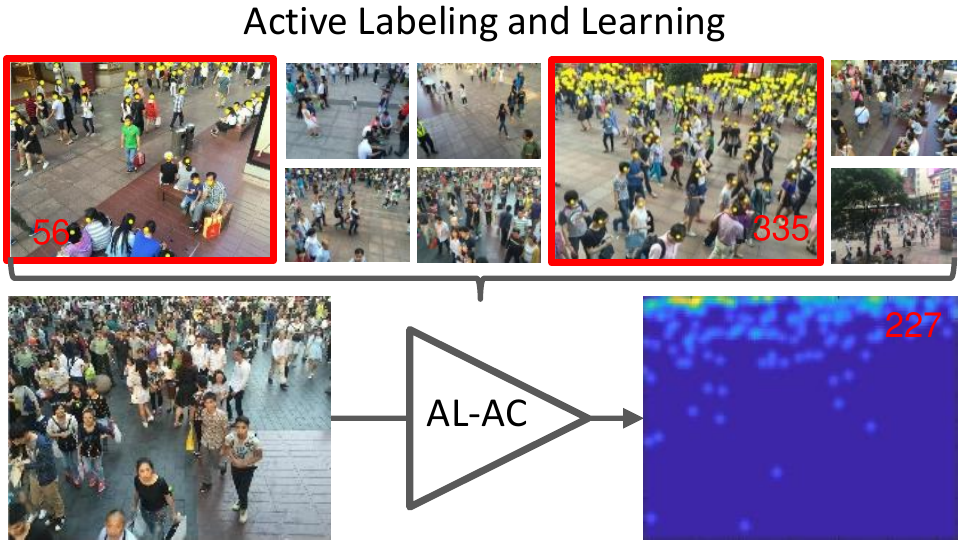}
	\caption{\small Given a crowd counting dataset, we propose an active learning framework (AL-AC) which actively labels only a small proportion of the dataset and learns an accurate density estimation network using both labeled and unlabeled data.
	}
	\label{Fig:motivation}
	\vspace{-0.5cm}
\end{figure}

Despite above, annotating head centers in dense crowds is still a laborious and tedious process. For instance, it can take up to 10 minutes for our annotators to annotate a single image with 500 persons; while the popular counting dataset ShanghaiTech PartA~\cite{zhang2016cvpr} has 300 training images with an average of 501 persons per image! To substantially reduce the annotation cost, we study the crowd density estimation in a semi-supervised setting where only handful images are labeled while the rest are unlabeled. This setting has not been largely explored in crowd counting: \cite{change2013cvpr,zhou2018tcsvt} propose to actively annotate the most informative video frames for semi-supervised crowd counting, yet the algorithms are not deep learning based and rely on frame consecutiveness. Recently, some deep learning works propose to leverage additional web data~\cite{liu2018bcvpr,liu2019tpami} or synthetic data~\cite{wang2019cvpr} for crowd counting; images in existing dataset are still assumed annotated, or at least many of them. The model transferability is also evaluated in some works~\cite{hossainone2019bmvc,xu2019iccv} where a network is trained on a source dataset with full annotations and tested on a target dataset with no/few annotations.

Given an existing dataset and a power DNN, we find that 1) learning from only a small subset,
the performance can vary a lot depending on the subset selection;
2) for specific subset that covers diverse crowd densities, the performance can be quite good (see results in Sec.~\ref{Sec:Ablation}).
This motivates us to study crowd counting with very limited annotations yet producing very competitive precision.
To achieve this goal,  we propose an Active Learning framework for Accurate crowd Counting (AL-AC) as illustrated in Fig.~\ref{Fig:motivation}:
given a labeling budget, instead of randomly selecting images to annotate,
we first introduce an active labelling strategy to iteratively
annotate the most informative images in the dataset and learn the counting model on them.
In each cycle we select samples that cover different crowd densities and also dissimilar to previous selections.
Eventually, the large amount of unlabeled data are also included into the network training:
we design a classifier with gradient reversal layer~\cite{ganin2014jmlr} to align the intrinsic distributions of labeled and unlabeled data.
Since all training samples contain the same object class, \eg person, we propose to further align distributions in-between training samples by mixing up the latent representations and distribution labels among labeled and unlabeled data in the network. With very limited labeled data, our model produces very competitive counting result.


To summarize, several new elements are offered:
\begin{compactitem}
	\item We introduce an active learning framework for accurate crowd counting with limited supervision.
	\item We propose a partition-based sample selection with weights (PSSW) strategy to actively select and annotate both diverse and dissimilar samples for network training.
	\item We design a distribution alignment branch with latent MixUp to align the distribution between the labeled data and large amount of unlabeled data in the network.
\end{compactitem}

Extensive experiments are conducted on standard counting benchmarks, \ie ShanghaiTech~\cite{zhang2016cvpr}, UCF\_CC\_50~\cite{idrees2013cvpr}, Mall~\cite{chen2012bmvc},TRANCOS~\cite{guerrero2015ibpra}, and DCC~\cite{marsden2018cvpr}. Results demonstrate that, with a small number of labeled data, our AL-AC reaches levels of performance not far from state of the art fully-supervised methods.

\section{Related works}\label{Sec:RelatedWorks}
In this section, we mainly survey deep learning based crowd counting methods and discuss semi-supervised learning and active learning in crowd counting. 

\subsection{Crowd counting}
The prevailed crowd counting solution is to estimate a density map of a crowd image, whose integral of the density map gives the total person count of that image~\cite{zhang2016cvpr}. A density map encodes spatial information of an image, regressing it in a DNN is demonstrated to be more robust than simply regressing a global crowd count~\cite{zhang2015cvpr,zhang2018wacv}. Due to the commonly occurred heavy occlusions and perspective distortions in crowd images, multi-scale or multi-resolution architectures are often exploited in DNNs: Ranjan~\etal \cite{ranjan2018eccv} propose an iterative crowd counting network
which produces the low-resolution density map and uses it to generate the high-resolution density map. Cao~\etal~\cite{cao2018eccv} propose a novel encoder-decoder network, where the encoder extracts multi-scale features with scale aggregation modules and the decoder generates high-resolution density
maps by using a set of transposed convolutions. Furthermore, Jiang~\etal~\cite{jiang2019cvpr} develop a trellis encoder-decoder network that incorporates multiple decoding paths to hierarchically aggregate features at different encoding stages.
In order to better utilize multi-scale features in the network, the attention~\cite{liu2018cvpr,shi2019iccv}, context~\cite{sindagi2017iccv,liu2019cvpr}, or perspective~\cite{shi2019cvpr,yan2019iccv} information in crowd images is often leveraged into the network.

Apart from density estimation based methods, some other variants in recent trends try to give the individual location or size information 
in crowd counting~\cite{liu2019cvprb,idrees2018eccv,laradji2018eccv,lian2019cvpr}. 
In order to achieve a good counting accuracy, 
they often integrate themselves into the density estimation pipeline. 
Our work is a density estimation based approach.

\subsection{Semi-supervised learning}
Semi-supervised learning~\cite{olivier2006tnn} refers to learning with a small amount of labeled data and a large amount of unlabeled data, and has been a popular paradigm in deep learning~\cite{weston2012nn,rasmus2015nips,laine2017iclr,yang2019arxiv}. It is traditionally studied for classification, where a label represents a class per image~\cite{lee2013icmlw,hoffer2016arxiv,rasmus2015nips,laine2017iclr}. In this work, we focus on semi-supervised learning in crowd counting, where the label of an image means the people count, with individual head points available in most cases. The common semi-supervised crowd counting solution is to leverage both labeled and unlabeled data into the learning procedure: Tan \etal~\cite{tan2011pr} propose a semi-supervised elastic net regression method by utilizing sequential information between unlabeled samples and their temporally neighboring samples as a regularization term; Loy~\etal~\cite{change2013cvpr} further improve it by utilizing both the spatial and temporal regularization in a semi-supervised kernel ridge regression problem; finally, in~\cite{zhou2018tcsvt}, graph Laplacian regularization and spatiotemporal constraints are incorporated into the semi-supervised regression. All these are not deep learning works and rely on temporal information among video frames.

Recently, Olmschenk~\etal~\cite{olmschenk2018wacv,olmschenk2019cviu} employ a generative adversarial network (GAN) in DNN to allow the usage of unlabeled data in crowd counting. Sam~\etal~\cite{sam2019aaai} introduce an almost unsupervised learning method that only a tiny proportion of model parameters is trained with labeled data while vast parameters are trained with unlabeled data. Liu~\etal~\cite{liu2018bcvpr,liu2019tpami} propose to learn from unlabeled crowd data via a self-supervised ranking loss in the network. In~\cite{liu2018bcvpr,liu2019tpami}, they mainly assume the existence of a labeled dataset and add extra data from the web; in contrast, our AL-AC seeks a solution for accurate crowd counting with limited labeled data. Our method is also similar to~\cite{olmschenk2018wacv,olmschenk2019cviu} in spirit of the distribution alignment between labeled and unlabeled data. While in~\cite{olmschenk2018wacv,olmschenk2019cviu} they need to generate fake images to learn the discriminator in GAN which makes it hard to learn and converge. Our AL-AC instead mixes the representations of labeled and unlabeled data in the network and learns the discriminator against them.
\medskip

\subsection{Active learning }
Active learning defines a strategy determining data samples that, when added to the training set, improve a previously trained model most effectively~\cite{sener2018iclr}. Although it is not possible to obtain an universally good active learning strategy~\cite{dasgupta2005nips},  there exist many heuristics~\cite{settles2009UWM}, which have been proved to be effective in practice. Active learning has been explored in many applications such as image classification~\cite{sinha2019iccv,joshi2009cvpr} and object detection~\cite{gonzalez2015cvpr}, while in this paper we focus on crowd counting.
Methods in this context normally assumes the availability of the whole counting set and choose samples from it, 
which is the so-called pool-based active learning~\cite{yang2015ijcv}.
\cite{change2013cvpr} and~\cite{zhou2018tcsvt} employ the graph-based approach to build adjacency matrix of all crowd images in the pool, 
sample selection is therefore cast as a matrix partitioning problem.
Our work is also pool-based active learning.

Lately, Liu~\etal~\cite{liu2019tpami} apply active learning in DNN where they measure the informativeness of unlabeled samples via mistakes made by the network on a self-supervised proxy task. The method is conducted iteratively and in each cycle it selects a group of images based their uncertainties to the model. The diversity of selected images is however not carefully taken care in their uncertainty measure, which might result in a biased selection within some specific count range. Our work instead interprets uncertainty from two perspectives: selected samples are diverse in crowd density and dissimilar to previous selection in each learning cycle. It should also be noted that \cite{liu2019tpami} mainly focuses on adding extra unlabeled data to an existing labeled dataset, 
while our AL-AC seeks for the limited data to be labeled within a given dataset. 

\section{Method}\label{Sec:Method}
\begin{figure*}[t]
	\centering
	\includegraphics[width=1\textwidth]{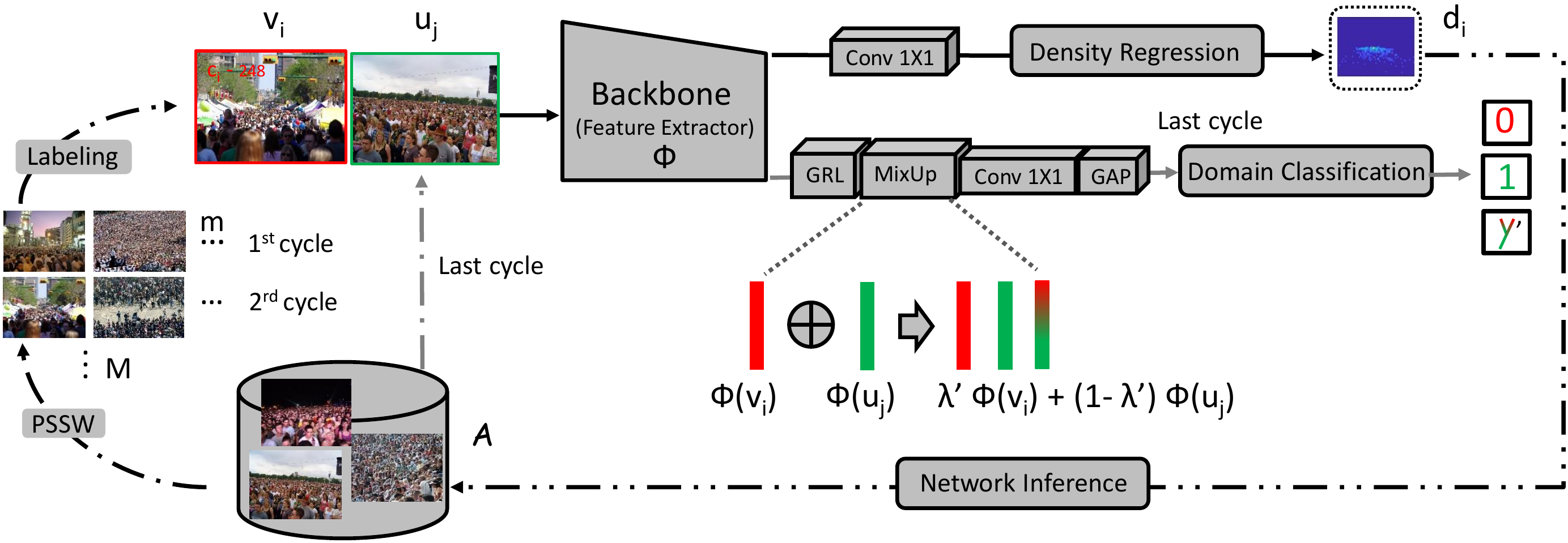}
	\caption{\small Overview of our active learning framework for accurate crowd counting. GRL: gradient reversal layer; GAP: global average pooling. PSSW: Partition-based  sample selection with weights; Conv 1$\times$1: output channel is 1.  Given a labeling budget $M$, we propose a PSSW strategy to actively select $m$ images to label in every cycle. Labeled samples $v_i$ are used to learn the density estimation model. We use the  model to do inference in the dataset $\mathcal A$ and repeat PSSW until the budget $M$ is met. In the last learning cycle, unlabeled data $u_j$ are also incorporated into network training via a proposed distribution alignment branch with latent MixUp. Notice MixUp applies between any two images, labeled or unlabeled; this figure only illustrates the case between one labeled and one unlabeled.
	}
	\label{Fig:intro}
	\vspace{-0.2cm}
\end{figure*}
\subsection{Problem}\label{Sec:problem}

We follow \emph{crowd density estimation} in deep learning context where density maps are pixel-wise regressed in a DNN~\cite{zhang2016cvpr,li2018cvpr}. A ground truth density map is generated by convolving Gaussian kernels at head centers in an image~\cite{zhang2016cvpr}. The network is optimized through a loss function minimizing the prediction error over the ground truth. In this paper, we place our problem in a \emph{semi-supervised} setting where we only label several or few dozens of images  while the rest large amount remains unlabeled. Both the labeled and unlabeled data will be exploited in model learning. Below, we introduce our active learning framework for accurate crowd counting (AL-AC).

\subsection{Overview} \label{Sec:Overview}
Our algorithm follows an active learning pipeline in general. It is an iterative process where a model is learnt in each cycle and a set of samples is chosen to be labeled from a pool of unlabeled samples~\cite{settles2009UWM}. In classic setting, only one single sample is chosen in each cycle. This is however not feasible for DNNs because
it is infeasible to train as many models as the number of samples since many practical problems of interest are very large-scale~\cite{sener2018iclr}. Hence, the commonly used strategy is batch mode selection~\cite{wang2016tcsvt,liu2019tpami} where a subset is selected and labeled in each cycle. This subset is added into the labeled set to update the model and repeat the selection in next cycle. The procedure continues until a predefined criterion is met, \eg a fixed budget.

Our method is illustrated in Fig.~\ref{Fig:intro}: given a dataset $\mathcal A$ with labeling budget $M$ (number of images as in~\cite{sam2019aaai,liu2019tpami}), we start by labeling $m$ samples uniformly at random from $\mathcal A$. For each labeled sample $v_i$, we generate its count label $c_i$ and density map $d_i$ based on the annotated head points in $v_i$.
We denote $\mathcal V^1= \{v_i, c_i, d_i\}$ and $\mathcal U^1 = \{u_j\}$ as the labeled and unlabeled set in cycle 1, respectively. A DNN regressor $R^1$ is trained on $\mathcal V^1$ for crowd density estimation.  Based on $R^1$'s estimation of density maps on $\mathcal U^1$, we propose a partition-based sample selection with weights strategy to select and annotate $m$ samples from $\mathcal U^1$. These samples are added to $\mathcal V^1$ so we have the updated labeled and unlabeled set $\mathcal V^2$ and $\mathcal U^2$ in cycle 2. Model $R^1$ is further trained on $\mathcal V^2$ and updated as $R^2$. The prediction of $R^2$ is better than $R^1$ as it uses more labeled data, we use the new prediction on $\mathcal U^2$ to again select $m$ samples and add them to $\mathcal V^2$. The process moves on until the labeling budget $M$ is met.
The unlabeled set $\mathcal U$ is also employed in network training through our proposed distribution alignment with latent MixUp. We only use $\mathcal U$ ($\mathcal U^T$) in the last learning cycle $T$ as we observe that adding it in every cycle does not bring us accumulative benefits but rather additional training cost.

The backbone network is not specified in Fig.~\ref{Fig:intro} as it can be any standard backbone. We will detail our selection of backbone, $M$, $m$ and $R$ in Sec.~\ref{Sec:Experiment}. Below we introduce our partition-based sample selection with weights and distribution alignment with latent MixUp. Overall loss function is given in this end.

\subsection{Partition-based sample selection with weights (PSSW)} \label{sec:pwrs}
In each learning cycle, we want to annotate the most informative/uncertain samples and add them to the network. The \emph{informativeness/uncertainty} of samples is evaluated from two perspectives: \emph{diverse} in density and \emph{dissimilar} to previous selections. It is observed that crowd data often forms a well structured manifold where different crowd densities normally distribute smoothly within the manifold space~\cite{change2013cvpr}; the {diversity} is to select crowd samples that cover different crowd densities in the manifold. This is realized by separating the unlabeled set into different density partitions for diverse selection.
Within each partition, we want to select those samples that are dissimilar to previous labeled samples, such that the model has not seen them. The dissimilarity is measured considering both local crowd density and global crowd count: we introduce a grid-based dissimilarity measure (GDSIM) for this purpose. Below, we formulate our partition-based sample selection with weights.

\begin{figure}[t]
	\centering
	\includegraphics[width=1\textwidth]{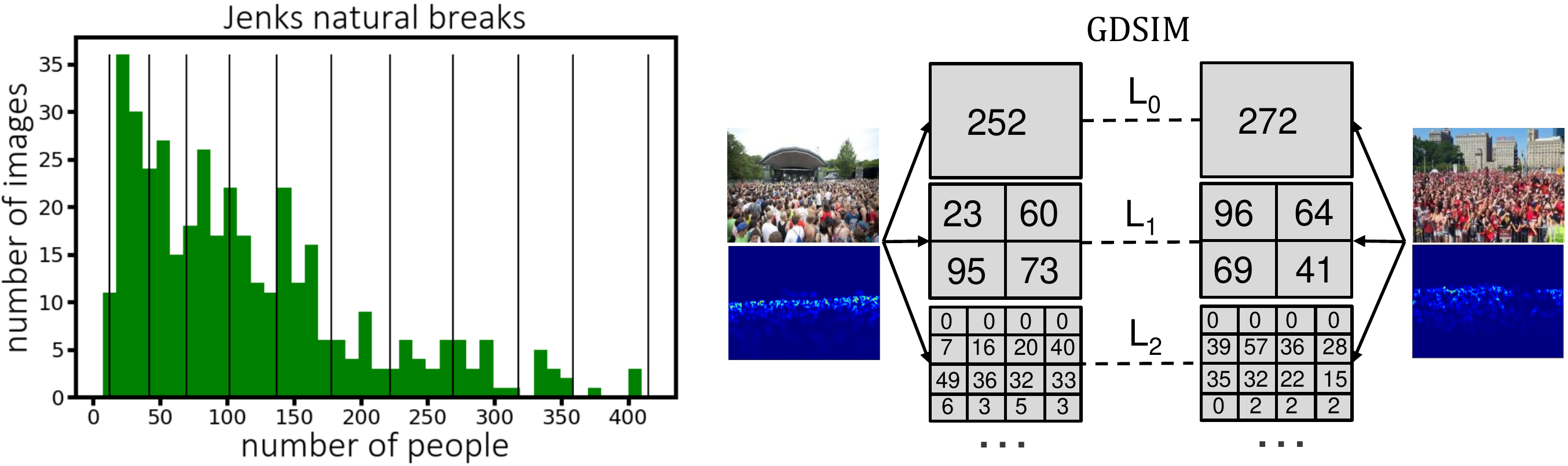}
	\caption{\small Illustration of Jenks natural breaks (Left) and grid-based dissimilarity measure (GDSIM, Right). We take the histogram of crowd count on SHB.
	}
	\label{Fig:gdsim}
	\vspace{-0.3cm}
\end{figure}

Formally, given the model $R^t$, unlabeled set $\mathcal U^t$ and labeled set $\mathcal V^t$ in $t^{th}$ cycle, we denote by  $\widetilde c_j$ the predicted crowd count by $R^t$ for an unlabeled image $u_j$. The histogram of all $\widetilde c_j$ on $\mathcal U^t$ discloses the overall density distribution. For the sake of diversity, we want to partition the histogram into $m$ parts and select one sample from each. Since the crowd counts are not evenly distributed (see Fig.~\ref{Fig:gdsim}: Left), sampling images evenly from the histogram
can end up with a biased view of the original distribution. We therefore employ the Jenks natural breaks optimization~\cite{jenks1967data} to partition the histogram. Jenks minimizes the variation within each range, so the partitions between ranges reflect the natural breaks of the histogram (Fig.~\ref{Fig:gdsim}).

Within each partition $P_k$, inspired by grid average mean absolute error (GAME)~\cite{guerrero2015ibpra}, we propose a grid-based dissimilarity from an unlabeled sample to labeled samples. Given an image $i$, GAME is originally introduced as an evaluation measure for density estimation,
\begin{equation}\label{Eq:game}
\text{GAME}(L) = \sum_{l=1}^{4^L}|\widetilde{c_i^{\hspace{.02in}l}} - c_i^{\hspace{.02in}l}|,
\end{equation}
where $\widetilde{c_i^{\hspace{.02in}l}}$ is the estimated count in region $l$ of image $i$. It can be obtained via the integration over the density $\widetilde{d_i^{\hspace{.02in}l}}$ of that region $l$; $c_i^{\hspace{.02in}l}$ is the corresponding ground truth count. Given a specific level $L$, GAME$(L)$ subdivides the image using a grid of $4^L$ non-overlaping regions which cover the full image (Fig.~\ref{Fig:gdsim}); the difference between the prediction and ground truth is the sum of the mean absolute error (MAE) in each of these regions. With different $L$, GAME indeed offers moderate ways to compute the dissimilarity between two density maps, taking care of both global counts and local details. Building on GAME, we introduce grid-based dissimilarity measure GDSIM as,
\begin{equation}\label{Eq:dsim}
\mathop {{\text{GDSIM}(u_j, L_A)}}\limits_{{u_j\in \mathcal P_k}}  = \min_{i, v_i \in \mathcal P_k} \bigg (\sum_{L=0}^{L_\text{A}}\sum_{l=1}^{4^{L}}|\widetilde{c_j^{\hspace{.02in}l}} - c_i^{\hspace{.02in}l}|\bigg),
\end{equation}
where $u_j$ and $v_i$ are from the unlabeled set $\mathcal U^t$ and labeled set $\mathcal V^t$, respectively;
they both fall into the $\mathcal P_k$-th partition. $\widetilde{c_i^{\hspace{.02in}l}}$ and $c_i^l$ are crowd counts in region $l$ as in formula~(\ref{Eq:game}) but for different images $u_j$ and $v_i$ (see Fig.~\ref{Fig:gdsim}: Right). Given the level $L_A$, unlike GAME, we compute the dissimilarity between $u_j$ and $v_i$ by traversing all levels from $0$ to $L_A$ (Fig.~\ref{Fig:gdsim}). In this way, the dissimilarity is computed based on both global count ($L = 0$) and local density ($L = L_A$) differences. Afterwards, instead of averaging the dissimilarity scores from $u_j$ to all the $v_i$ in $\mathcal P_k$, we use $\min$ to indicate if $u_j$ is closer to any one of the labeled images, it is regarded as a familiar sample to the model. Ideally, we should choose the most dissimilar sample from each partition; nevertheless, the crowd count $\widetilde{c_j^{\hspace{.02in}l}}$ in formula (\ref{Eq:dsim}) is not ground truth. We convert the GDSIM scores to probabilities and adopt weighted random selection  to label one sample from each partition.


\subsection{Distribution alignment with latent MixUp}\label{Sec:MixUP}
Since labeled data only represents partial crowd manifold, particularly when they are limited, distribution alignment with large amount of unlabeled data becomes necessary even within the same domain.
In order for the model to learn a proper subspace representation of the entire set, we introduce  distribution alignment with latent MixUp.

We assign labeled data with distribution labels 0 while unlabeled data with labels 1. A distribution classifier branched off from the deep extractor ($\phi$ in Fig.~\ref{Fig:intro}) is designed: it is composed of a gradient reversal layer (GRL)~\cite{ganin2014jmlr}, 1 $\times$ 1 convolution layer and global average pooling (GAP) layer.
The GRL multiplies the gradient by a certain negative constant (-1 in this paper) during the network back propagation; it enforces that the feature distributions over the labeled and unlabeled data are made as indistinguishable as possible for the distribution classifier, thus aligning them together.

The hard distribution labels create hard boundaries between labeled and unlabeled data. To further merge the distributions and particularly align in-between training samples, we adapt an idea from MixUp~\cite{zhang2018iclr}. MixUp normally trains a model on random convex combinations of raw inputs and their corresponding labels. It encourages the model to behave linearly ``between" training samples, as this linear behavior reduces the amount of undesirable oscillations when predicting outside the training samples. It has been popularly employed in several semi-supervised classification works~\cite{berthelot2019arxiv,verma2019icml,verma2019arxiv,zhang2018iclr}.
In this work, we integrate it into our distribution alignment branch for semi-supervised crowd counting.
We find that mixing raw input images does not work for our problem. Instead we propose to mix their latent representations in the network:
supposedly we have two images, $x_1$, $x_2$, and their distribution labels $y_1$, $y_2$, respectively. The latent representations of $x_1$ and $x_2$ are produced by the deep extractor $\phi$ as two tensors ($\phi(x_1)$ and $\phi(x_2)$) from the last convolutional layer of the backbone. We mix up ($\phi(x_1)$, $y_1$), ($\phi(x_2)$, $y_2$) with a weight $\lambda'$ as
\begin{equation} \label{Eq:MixUp}
\begin{split}
& z' = \lambda'\phi(x_1) + (1-\lambda')\phi(x_2)\\
& y' = \lambda' \times y_1  + (1-\lambda') \times y_2.
\end{split}
\end{equation}
where ($z'$, $y'$) denotes the mixed latent representation and label. $\lambda'$ is generated in the same way with~\cite{berthelot2019arxiv}: $\lambda' = max(\lambda, 1-\lambda)$, $\lambda \sim \text{Beta}(\alpha, \alpha)$;
$\alpha$ is a hyper-parameter set to 0.5. {Both labeled and unlabeled data can be mixed.} For two samples with the same label, their mixed label remains. We balance the number of labeled and unlabeled data with data augmentation (see Sec.~\ref{Sec:Dataset}) so a mixed pair can be composed of labeled or unlabeled data with (almost) the same probability.
MixUp enriches the distribution in-between training samples. Together with GRL, it allows the network to elaborately knit the distributions of labeled and unlabeled data. The alignment is only carried out in the last active learning cycle as an efficient practice. The network training proceeds with a multi-task optimization that minimizes the density regression loss on labeled data and the distribution classification loss for all data including mixed ones, specified below.

\subsection{Loss function}
For density regression task, we adopt the commonly used pixel-wise MSE loss $\mathcal L_{reg}$:
\begin{equation}
\mathcal L_{reg} = \frac{1}{2K}\sum_{k=1}^K \|d_k^{\hspace{.01in}e} - d_k^{\hspace{.01in}g}\|_2^2
\end{equation}
$d_k^{\hspace{.01in}e}$ and $d_k^{\hspace{.01in}g}$ denote the density map prediction and ground truth of image $k$, respectively. $K$ is the number of labeled images.
For the distribution classification task, since distribution labels for mixed samples can be non-integers, we adopt the binary cross entropy with logits loss $\mathcal L_{dc}$, which combines a Sigmoid layer with the binary cross entropy loss.  Given an image pair, $\mathcal L_{dc}$ is computed on each individual as well as their mixed representations (see Fig.~\ref{Fig:intro}).
The overall multi-task loss function is given by
\begin{equation}\label{Eq:Loss}
\mathcal L = \mathcal L_{reg} + \beta\mathcal L_{dc}
\end{equation}



\section{Experiments}\label{Sec:Experiment}
We begin by introducing five counting datasets: ShanghaiTech~\cite{zhang2016cvpr}, UCF\_CC\_50~\cite{idrees2013cvpr}, Mall~\cite{chen2012bmvc}, TRANCOS~\cite{guerrero2015ibpra}, and DCC~\cite{marsden2018cvpr}. It covers people~\cite{zhang2016cvpr,idrees2013cvpr,chen2012bmvc}, vehicle~\cite{guerrero2015ibpra} and cell~\cite{marsden2018cvpr} counting to demonstrate the generalization ability of our method.  
\subsection{Experimental Setup}\label{Sec:Dataset}

\para{Datasets.} \emph{ShanghaiTech}~\cite{zhang2016cvpr} consists of 1,198 annotated images with a total of 330,165 people with head center annotations. This dataset is split into SHA and SHB. The average crowd counts are 123.6 and 501.4, respectively. Following~\cite{zhang2016cvpr}, we use 300 images for training and 182 images for testing in SHA; 400 images for training and 316 images for testing in SHB.
\emph{UCF\_CC\_50}~\cite{idrees2013cvpr} has 50 images with 63,974 head center annotations in total. The head counts range between 94 and 4,543 per image. The small dataset size and large variance make this a very challenging counting dataset.  We call it UCF for short. Following~\cite{idrees2013cvpr}, we perform 5-fold cross validations to report the average test performance.
\emph{Mall}~\cite{chen2012bmvc} contains 2000 frames collected in a shopping mall. Each frame on average has only 31 persons. The first 800 frames are used as the training set and the rest 1200 frames as the test set.
\emph{TRANCOS}~\cite{guerrero2015ibpra} is a public traffic dataset containing 1244 images of different congested traffic scenes captured by surveillance cameras with 46,796 annotated vehicles. 800 images are for training and the rest for testing.
\emph{DCC}~\cite{marsden2018cvpr} is a cell microscopy dataset, consisting of 177 images with a cell count from 0 to 100. 100 images are used for training and 77 images are used for testing.

\medskip

\para{Implementation details.}\label{Sec:ExperimentalDetails}
The backbone ($\phi$) design follows~\cite{li2018cvpr}: VGGnet with 10 convolutional and 6 dilated convolutional layers, it is pretrained on ILSVRC classification task.
We follow the setting in~\cite{li2018cvpr} to generate ground truth density maps.
To have a strong baseline, the training set is augmented by randomly cropping patches of 1/4 size of each image.
We set a reference number 1200, both labeled and unlabeled data in each dataset are augmented up to this number to have a balanced distribution.
For instance, if we have 30 labeled images, we need to crop 40 patches from each image to augment it to 1200. We feed the network with a minibatch of two image patches each time. In order to have the same size of two patches, we further crop them to keep the shorter width and height of the two. We set the learning rate as 1e-7, momentum 0.95 and weight decay 5e-4. We train 100 epochs with SGD optimizer for each active learning cycle and before the last cycle, the network is trained with only labeled data. In the last cycle, it is trained with both labeled and unlabeled data. In all experiments, $L_A$ is 3 for GDSIM (\ref{Eq:dsim}) and $\beta$ is 3 for loss weight (\ref{Eq:Loss}).
Network inference is on the entire image.

\medskip


\para{Evaluation protocol.}\label{Sec:Protocol}
We evaluate the counting performance via the commonly used mean absolute error (MAE) and mean square error (MSE)~\cite{sam2017arxiv,sindagi2017iccv,liu2018cvpr} which measures the difference between the counts of ground truth and estimation. For active learning,
we choose to label around 10\% images of the entire set, which goes along with our setting of limited supervision. $m$ is chosen not too small so that we can normally reach the labeling budget in about 2-4 active learning cycles. Sec.~\ref{Sec:dis} gives a discussion on the time complexity. $M$ and $m$ are by default 30/40 and 10 on SHA and SHB, 10 and 3 on UCF (initial number is 4), 80 and 20 on Mall and TRANCOS, 10 and 3 on DCC, respectively. We also evaluate different $M$ and $m$ to show the effectiveness of our method.
The baseline is to randomly label $M$ images and train a regression model using the same backbone with our AL-AC but without distribution alignment. As in~\cite{change2013cvpr,zhou2018tcsvt}, taken the randomness into account, we repeat each experiment with 10 trials for both mean and standard deviation, to show the improvement of our method over baseline.


\subsection{ShanghaiTech}\label{Sec:Ablation}
We present ablation study of our AL-AC and its comparison to state of the art fully- and semi-supervised methods.

\para{Ablation study.} The proposed partition-based sample selection with weights and distribution alignment with latent MixUp are ablated.

\noindent \textbf{\emph{Labeling budget $M$ and $m$.}}  As mentioned in Sec.~\ref{Sec:Protocol}, we set $M= 30/40$ and $m= 10$ by default. Comparable experiments are offered in two ways. First, keeping $m = 10$, we vary $M$ from 10 to 40. The results are shown in Table~\ref{Tab:LabellingBudgetA}. We compare our  partition-based sample selection with weights (PSSW) with random selection (RS); distribution alignment is not added in this experiment. For PSSW, its MAE on SHA is gradually decreased from 121.2 with $M = 10$ to 85.4 with $M = 40$, the standard deviation is also decreased from 9.3 to 2.5. The MAE result is in general 10 points lower than RS. With different $M$, PSSW also produces lower MAE than RS on SHB. For example, with $M = 40$, PSSW yields an MAE of 14.6 \vs 17.9 for RS.

\begin{table}[t]
	\setlength{\tabcolsep}{3.3pt}
	\centering
	\scriptsize
	\begin{tabular}{ccccc}
		\toprule
		Dataset  & \multicolumn{2}{c}{SHA}  & \multicolumn{2}{c}{SHB}\\
		\midrule
		Method & PSSW & RS & PSSW & RS \\
	\midrule
		M= 10, m=10  &  121.2 \pms{9.3} & 121.2 \pms{9.3}  & 20.5 \pms 4.8 & 20.5 \pms 4.8 \\
		M=20, m=10 &  96.7\pms{7.3} & 111.5 \pms{7.4} & 17.0 \pms 1.9 & 19.3 \pms 2.2 \\
		M=30, m=10   & 93.5\pms{2.9} & 102.1 \pms{7.0}  & 15.7 \pms 1.5 & 19.9 \pms 3.1 \\
		M=40, m=10  & \textbf{85.4 \pms{2.5}} & \textbf{93.8 \pms{5.6}}  &\textbf{14.6 \pms 1.3} & \textbf{17.9 \pms 1.9} \\
		\midrule
		M = 30, m = 5  & 92.6 \pms{3.1} & 102.1 \pms{7.0}   & 15.1 \pms 1.5 & 19.9 \pms 3.1 \\
		M = 40, m = 5  & \textbf{84.4 \pms{2.6}} & \textbf{93.8 \pms{5.6}} & \textbf{14.4 \pms 1.2}   & \textbf{17.9 \pms 1.9}  \\
		\bottomrule
	\end{tabular}
		\hspace{0.5cm}
		\begin{tabular}{ccc}
			\toprule
			M=40, m=10   & SHA & SHB  \\
			\midrule
		RS (Baseline) & 93.8 & 17.9\\
			Even Partition & 89.6  & 16.2\\
			Global Diff & 86.6 &  15.3  \\
			PSSW &  \textbf{84.4}  &  \textbf{14.4}\\
		\bottomrule
		\end{tabular}
	\caption{\small Ablation study of the proposed partition-based sample selection with weights (PSSW) strategy. Left: comparison against random selection (RS). Right: comparison to some variants of PSSW; Even Partition means evenly splitting on the histogram of crowd count; Global Diff refers to using global count difference for dissimilarity.  MAE is reported on SHA and SHB. }
	\label{Tab:LabellingBudgetA}	
	\vspace{-0.6cm}
\end{table}

\begin{table}[t]
	\setlength{\tabcolsep}{3pt}
	\centering
	\scriptsize
	\begin{tabular}{ccccc}
		\toprule
		Dataset  & \multicolumn{2}{c}{SHA} & \multicolumn{2}{c}{SHB} \\
		\midrule
		M = 30, m =10 & MAE & MSE & MAE & MSE \\
	\midrule
		PSSW  &  93.5 \pms 2.9  & 151.0 \pms 15.1 & 15.7 \pms 1.5 &  28.3 \pms 3.4\\
		PSSW + GRL &  90.8 \pms 2.7  & 144.9 \pms 14.5  & 14.7 \pms 1.3  & 27.8 \pms 2.9 \\
		PSSw + GRL + MX     &  \textbf{87.9 \pms 2.3}  & \textbf{139.5 \pms 12.7}  & \textbf{13.9 \pms 1.2}  & \textbf{26.2 \pms 2.5}  \\
 \bottomrule
 \toprule
		M = 40, m =10 & MAE & MSE & MAE & MSE \\
	\midrule
		PSSW   &  85.4 \pms 2.5  & 144.7 \pms 10.7 & 14.6 \pms 1.3 &  24.6 \pms 3.0\\
		PSSW + GRL & 82.7 \pms 2.4  & 140.9 \pms 11.3  & 13.7 \pms 1.3  & 23.5\pms 2.2 \\
		PSSW + GRL + MX     &  \textbf{80.4 \pms 2.4}   & \textbf{138.8 \pms 10.1}   & \textbf{12.7 \pms 1.1}  & \textbf{20.4 \pms 2.1}  \\
		\bottomrule
	\end{tabular}
	\hspace{0.5cm}
		\begin{tabular}{ccc}
			\toprule
			M=40, m=10   & SHA & SHB  \\
			\midrule
			RS (Baseline) & {93.8 }  & 17.9 \\
				RS+ GRL + MX & 87.3  &  15.1 \\
				PSSW & 84.4  &  {14.4 }\\
					PSSW + GRL + MX &  \textbf{80.4 }   &  \textbf{12.7}\\
		\bottomrule
		\end{tabular}
	\caption{\small  Ablation study of the proposed distribution alignment with latent MixUp. Left: analysis on latent MixUp (MX) and gradient reversal layer (GRL). Right: comparison against RS plus GRL and MX. MAE is reported in the right table.  }
	\label{Tab:LabellingCycle}	
		\vspace{-0.7cm}
\end{table}


Second, by keeping $M = 30/40$, we decrease $m$ from 10 to 5 and repeat the experiment. Results show that having a small $m$ indeed works slightly better: for instance, PSSW with $M = 30$ and $m = 5$ reduces MAE by 1.0 on SHA compared to PSSW with $M = 30$ and $m = 10$. On the other hand, $m$ can not be too small as discussed in Sec.~\ref{Sec:Overview} and Sec.~\ref{Sec:dis}. In practice, we still keep $m = 10$ for both efficiency and effectiveness.

We notice that the number of labeled people may vary over trials and cycles. Since we do not know the ground truth, we can not make the number of labeled people exactly what we want before labeling them. As in~\cite{liu2019tpami,zhou2018tcsvt,change2013cvpr}, the essential idea of active learning based crowd counting is to find the most informative images to label within a small budget of number of images. Labelling more heads or less does not mean a better or worse performance. To give an insight, we conduct an experiment by only labeling images with over 200 heads on SHB, the MAE is 26.7 \vs 17.9 for RS in Table~\ref{Tab:LabellingBudgetA}.


\noindent  \textbf{\emph{Variants of PSSW.}} Our PSSW has two components: the Jenks-based partition for diversity, and the GDSIM for dissimilarity (Sec.~\ref{Sec:Method}). In order to show the effectiveness of each, we present two variants of PSSW: Even Partition and Global Diff.
Even Partition means that Jenks-based partition is replaced by evenly splitting the ranges on the histogram of crowd count while GSDIM remains; Global Diff means that GDSIM is replaced by using the global count difference to measure the dissimilarity while Jenks-based partition remains. We report MAE on SHA and SHB in Table~\ref{Tab:LabellingBudgetA}: Right. It can be seen that Even Partition produces MAE 89.6 on SHA and 16.2 on SHB, while Global Diff produces 86.6 and 15.3. Both are clearly inferior to PSSW (84.4 and 14.4). This suggests the importance of the proposed diversity and dissimilarity measure.

\noindent  \textbf{\emph{Distribution alignment with latent MixUp.}}
Our proposed distribution alignment with latent MixUp is composed of two elements: distribution classifer with GRL and latent MixUp (Sec.~\ref{Sec:MixUP}). To demonstrate their effectiveness, we present the result of PSSW plus GRL classifer (denoted as PSSW + GRL), and latent MixUp (denoted as PSSW + GRL + MX) in Table~\ref{Tab:LabellingCycle}.
We take $M = 40$ as an example, adding GRL and MX to PSSW contributes to 5.0 points MAE decrease on SHA and 1.9 points decrease on SHB. 
Specifically, The MX contributes to 2.3 and 1.0 points decrease on SHA and SHB, respectively. The same observation goes for MSE: by adding GRL and MX, it decreases from 144.7 to 138.8 on SHA, from 24.6 to 20.4 on SHB.

To make a further comparison, we also add the proposed distribution alignment with latent MixUp to RS in Table~\ref{Tab:LabellingCycle}: Right, where we achieve MAE 87.3 on SHA and 15.1 on SHB. Adding GRL+MX to RS also improves the baseline: the performance difference between PSSW and RS becomes smaller; yet, the absolute value of the difference is still big, which justifies our PSSW. Notice PSSW + GRL + MX is the final version of our AL-AC, we use AL-AC hereafter to denote it.


%


\begin{figure}[t]
	\centering
	\includegraphics[width=0.65\columnwidth]{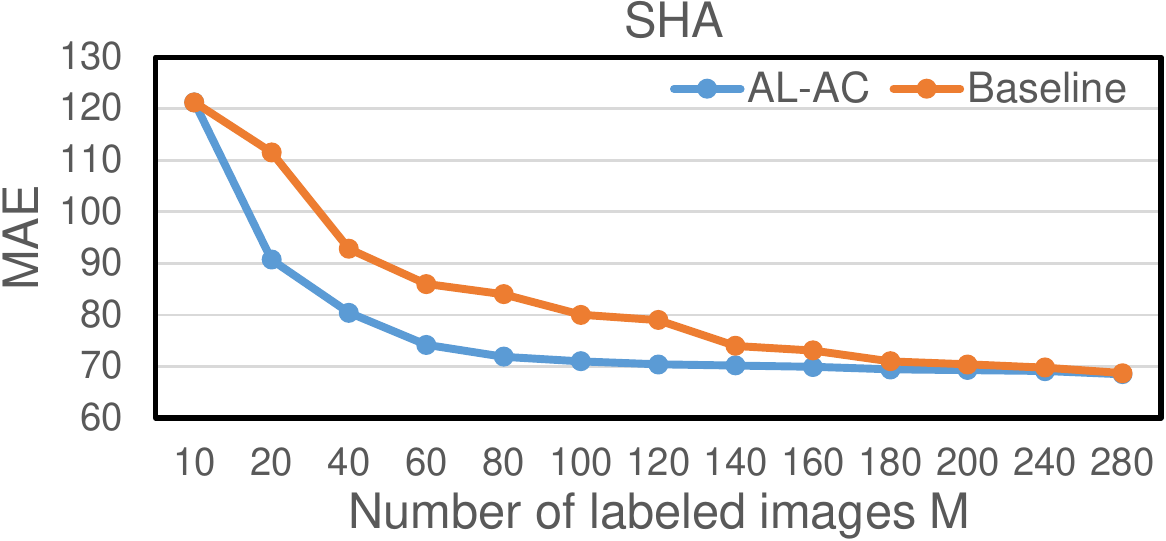}
	\caption{\small Comparison of AL-AC against baseline. MAE is reported on SHA. The labeling budget $M$ increases from 10 to 280. }
	\label{Fig:curve}
	\vspace{-0.4cm}
\end{figure}

\para{Comparison with fully-supervised methods.} We compare our work with those prior arts~\cite{zhang2016cvpr,sam2017arxiv,li2018cvpr,ranjan2018eccv,shi2019cvpr,shi2019iccv,ma2019iccv}. All these approaches are fully-supervised methods which utilize annotations of the entire dataset (300 in SHA and 400 in SHB). While in our setting, we label only 30/40 images, 10\% of the entire set. It can be seen that our method outperforms the representative methods~\cite{zhang2016cvpr,sam2017arxiv} a few years ago, and are not far from other recent arts, \ie~\cite{li2018cvpr,ranjan2018eccv,shi2019cvpr,shi2019iccv,ma2019iccv}. A direct comparison to ours is CSRNet~\cite{li2018cvpr}, we share the same backbone. With about 10\% labeled data, our AL-AC retains 85\% accuracy on SHA (68.2 / 80.4), 83\% accuracy on SHB (10.6 / 12.7 ). Compared to our baseline (denoted as RS in Table~\ref{Tab:LabellingBudgetA}), AL-AC in general produces significantly lower MAE, \eg 87.9 \vs 102.1 on SHA with $M = 30$;  17.9 \vs 12.7 on SHB with $M = 40$.

Despite that we only label 10\% data, our distribution alignment with latent MixUp indeed enables us to make use of more unlabeled data across datasets: for instance, a simple implementation with M = 40 on SHA, if we add SHB as unlabeled data to AL-AC for distribution alignment, we obtain an even lower MAE 78.6 \vs 80.4 in Table~\ref{Tab:Counting}.

\para{Comparison with semi-supervised methods.} There are also some semi-supervised crowd counting methods~\cite{liu2019tpami,sam2019aaai,olmschenk2019cviu}\footnote{Results of ~\cite{liu2019tpami,sam2019aaai} can be estimated from their curve plots.}. For instance in \cite{sam2019aaai,olmschenk2019cviu}, with $M = 50$ they produce MAE 170.0 and 136.9 on SHA, respectively. These are much higher MAE than ours. Since \cite{sam2019aaai,olmschenk2019cviu} use different architectures from AL-AC, they are not straightforward comparisons. For~\cite{liu2019tpami}, it uses about 50\% labeled data on SHA (Fig.7 in~\cite{liu2019tpami}) to reach the similar performance of our AL-AC with 10\% labeled data. We both adopt the VGGnet yet \cite{liu2019tpami} utilizes extra web data for ranking loss while we only use unlabeled data within SHA, we use dilated convolutions while~\cite{liu2019tpami} does not. To make them more comparable, we instead use the same backbone of~\cite{liu2019tpami} and repeat AL-AC on SHA (implementation details still follow Sec.~\ref{Sec:Dataset}), the mean MAE with M=30, m=10 on SHA becomes 91.4 (\vs 87.9 in Table~\ref{Tab:Counting}), which is still much better than that of~\cite{liu2019tpami}.

Instead of using limited labeled data, in Fig.~\ref{Fig:curve}, we keep increasing $M$ till 280 and report the MAE on SHA. It can be seen that, with about 80-100 labeled data (nearly 30\%) labeled data, AL-AC already reaches the performance close to the fully-supervised method, as in~\cite{li2018cvpr} (Table~\ref{Tab:Counting}). The performance will saturate after some point and converge to that
of baseline. This is also observed in other works~\cite{liu2019tpami,sam2019aaai}.

\begin{table}[!t]
\begin{minipage}[!t]{0.45\columnwidth}
  \renewcommand{\arraystretch}{1.3}
  \centering
  \scriptsize
  \setlength{\tabcolsep}{0.8mm}{
  	\begin{tabular}{ccccc}
			\toprule
		Dataset  & \multicolumn{2}{c}{SHA}& \multicolumn{2}{c}{SHB} \\
\cmidrule{2-5}
		Measures & MAE& MSE & MAE & MSE  \\
	   \midrule
		MCNN~\cite{zhang2016cvpr} & 110.2 & 173.2 & 26.4 & 41.3  \\
		Switching CNN~\cite{sam2017arxiv}  & 90.4 & 135.0 & 21.6 & 33.4 \\
		CSRNet \cite{li2018cvpr} & {68.2} & 115.0 & {10.6} & {16.0}\\
		ic-CNN~\cite{ranjan2018eccv}& 68.5 & 116.2 & 10.7 & 16.0 \\
		PACNN~\cite{shi2019cvpr} & \textbf{62.4} & {102.0} & 7.6& 11.8 \\
		CFF~\cite{shi2019iccv} & 65.2 & 109.4 & \textbf{7.2} & {11.2} \\
        BAYESIAN+~\cite{ma2019iccv} & 62.8 & \textbf{101.8} & 7.7& \textbf{12.7} \\
	\midrule
	Baseline (M = 30)& {102.1} & {164.0} & {19.9} & {30.6} \\
		AL-AC (M = 30)  & \textbf{87.9} & \textbf{139.5} & \textbf{13.9} & \textbf{26.2} \\
		Baseline (M =40) & {93.8} & {150.9} & {17.9} & {27.3} \\
		AL-AC (M =40) & \textbf{80.4} & \textbf{138.8} & \textbf{12.7} & \textbf{20.4} \\
		\bottomrule
	\end{tabular}
 \caption{\small Comparison of AL-AC to \\the state of the art on SHA and SHB. }
  \label{Tab:Counting}
	}
  \end{minipage}
\begin{minipage}[!t]{0.49\columnwidth}
  \renewcommand{\arraystretch}{1.3}
  \centering
   \scriptsize
  \setlength{\tabcolsep}{0.8mm}{
	\begin{tabular}{ccc}
			\toprule
		Counting & \multicolumn{2}{c}{UCF} \\
		\cmidrule{2-3}
		Measures & MAE& MSE  \\
				\midrule
	    MCNN~\cite{zhang2016cvpr} & 377.6 & 509.1 \\
	    Switching CNN~\cite{sam2017arxiv} & 318.1 & 439.2\\
	    	CP-CNN\cite{sindagi2017iccv} & 295.8 &  320.9\\
		CSRNet \cite{li2018cvpr}  & {266.1} &  {397.5} \\
		ic-CNN~\cite{ranjan2018eccv} & 260.0 & 365.5  \\
		PACNN~\cite{shi2019cvpr} & {241.7} & {320.7} \\
		BAYESIAN+~\cite{ma2019iccv} & \textbf{229.3} & \textbf{308.2} \\
		\midrule
				Baseline (M=10, m=3) & 444.7 \pms 25.9 &  600.3 \pms 32.7  \\
				AL-AC (M=10, m=3) & 351.4 \pms 19.2 & {448.1 \pms 24.5} \\
			Baseline (M=20, m=10) & 417.2 \pms 29.8 & 550.1 \pms 25.5 \\
		 AL-AC (M=20, m=10) & \textbf{318.7 \pms 23.0} &  \textbf{421.6 \pms 24.1}  \\
			\bottomrule
	\end{tabular}
	 \caption{\small Comparison of AL-AC with state of the art on UCF. }
	\label{Tab:Counting-UCF}	
	}
  \end{minipage}
  \vspace{-0.4cm}
\end{table}




\begin{figure*}[t]
	\centering
	\includegraphics[width=1\textwidth]{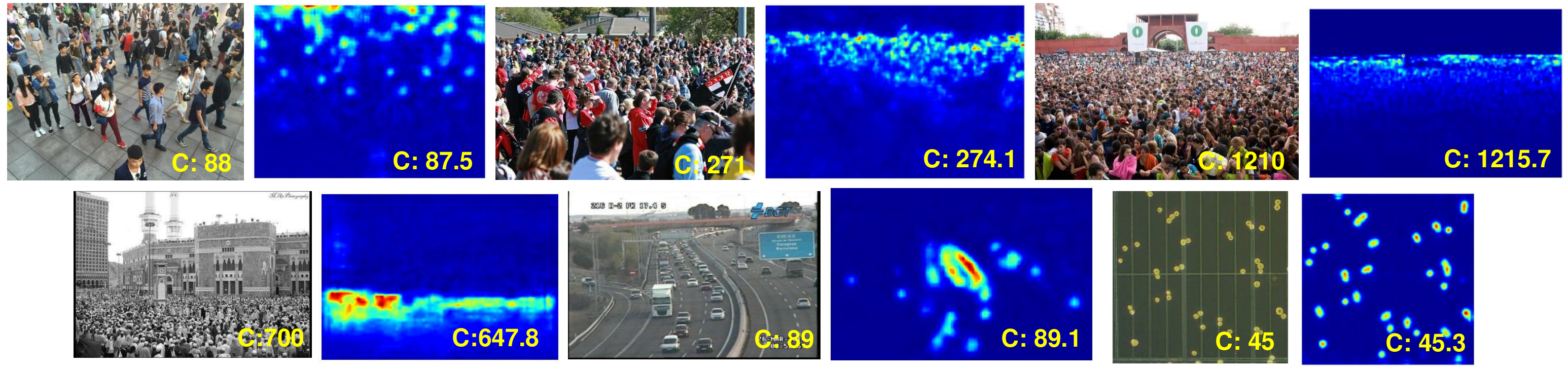}
	\caption{\small Examples of AL-AC on SHA, SHB, UCF, TRANCOS, and DCC datasets. Ground truth counts are  in the original images while predicted counts in the estimated  density maps. }
	\label{Fig:vislization}
	\vspace{-0.3cm}
\end{figure*}

\subsection{UCF\_CC\_50~~} It has 40 training images in total. We show in Table~\ref{Tab:Counting-UCF} that, labeling ten of them ($M = 10, m = 3$)
already produces a very competitive result: the MAE is 351.4 while the MSE is 448.1.
The MAE and MSE are significantly lower (93.3 and 152.2 points) than baseline. We analyzed the result and found that our AL-AC is able to select those hard samples with thousands of persons and label them for training, while this is not guaranteed in random selection. Compared to fully supervised method, \eg~\cite{li2018cvpr}, our MAE is  not far. We also present the result of $M = 20, m = 10$: MAE/MSE is further reduced.


\subsection{Mall~~} Different from ShanghaiTech and UCF datasets, Mall contains images with much sparser crowds, 31 persons on average per image. Following our setup, we label 80 out of 800 images and compare our AL-AC with both baseline and other fully-supervised methods~\cite{pham2015iccv,liu2018cvpr,hossain2019wacv} in Table~\ref{tab:mall}.
With 10\% labeled data, we achieve MAE 3.8 superior to the baseline and~\cite{pham2015iccv}, MSE 5.5 superior to the baseline and~\cite{xiong2017iccv}. This shows the effectiveness of our method on images of sparse crowds.

\begin{table}[t]
	\scriptsize
	\centering
	\setlength{\tabcolsep}{2.5pt}
	\begin{tabular}{c|cc|ccccc}
		\toprule
		Mall (M=80, m=20) & 	Baseline  &		AL-AC* & Count Forest~\cite{pham2015iccv} & ConvLSTM~\cite{xiong2017iccv}&	DecideNet~\cite{liu2018cvpr}& E3D~\cite{zou2019bmvc} &	SAAN~\cite{hossain2019wacv}  \\
		\midrule
		MAE & 5.9 \pms 0.9 &  3.8 \pms 0.5 & 4.4 & 2.1 & 1.5 & 1.6& 1.3  \\
		MSE & 6.3 \pms 1.1 & 5.4 \pms 0.8 & 2.4 & 7.6 & 1.9 & 2.1& 1.7\\
		\bottomrule
	\end{tabular}
		\caption{\small Comparison of AL-AC with state of the art on Mall. }
	\label{tab:mall}
		\vspace{-0.4cm}
\end{table}

\subsection{TRANCOS and DCC}
To test the generalization ability of our AL-AC on other counting tasks, we evaluate it on TRANCOS and DCC for vehicle and cell counting, respectively. The global count error MAE is presented in Table~\ref{tab:trancos}. We label 10\% of the images for each dataset. That is, $M = 80$, $m = 20$ for TRANCOS, and $M = 10$, $m = 3$ for DCC. Our MAE result is 7.5 on TRANCOS with an decrease of 2.6 points from baseline; 4.5 on DCC with an decrease of 2.9 points from baseline.  With 10\% labeled data, our AL-AC performs close to state of the art, particularly on DCC.

\begin{table}[t]
	\scriptsize
	\centering
	\setlength{\tabcolsep}{5.2pt}
	\begin{tabular}{c|cc|ccccc}
		\toprule
		\textsc{Method} & 	Baseline*  &		AL-AC* &  Lempitsky\cite{lempitsky2010nips} & Hydra-3s\cite{onoro2016eccv} & POCR~\cite{marsden2018cvpr} & 	 CSRNet~\cite{li2018cvpr}& 	CFF~\cite{shi2019iccv}  \\
		\midrule
		TRANCOS & 10.1 \pms 1.5 &  7.5 \pms 0.8 & 13.8 & 11.0 & 9.7 & 3.6 & 2.0  \\
		DCC    &  7.4 \pms 1.2 &  4.5 \pms 0.4 & - & - & 8.4 & -  & 3.2  \\
		\bottomrule
	\end{tabular}
	\caption{\small Comparison of AL-AC with state of the art on TRANCOS and DCC. Notice the labeling budget for TRANCOS and DCC is different. We label 10\% of images over each set, such that $M = 80$ and $m = 20$ for TRANCOS, and  $M = 10$ and $m = 3$ for DCC. MAE is reported. }
	\label{tab:trancos}
		\vspace{-0.5cm}
\end{table}



\section{Discussion}\label{Sec:dis}
We present an active learning framework for accurate crowd counting with limited supervision. Given a counting dataset, instead of exhaustively annotating every image, we first introduce a partition-based sample selection with weights to label only a few most informative images and learn a crowd regression network upon them. This process is iteratively repeated till the labeling budget is reached. Next, rather than learning from only labeled data, the abundant unlabeled data are also exploited: we introduce a  distribution alignment branch with latent MixUp in the network. Experiments conducted on standard benchmarks show that labeling only 10\% of the entire set, our method already performs close to recent state-of-the-art.

By choosing an appropriate $m$, we normally reach the labeling budget in three active learning cycles. In our setting, training data in each dataset are augmented to a fixed number. We run our experiments with GPU GTX1080. It takes around three hours to complete each active learning cycle. The total training hours are more or less the same to fully-supervised training, as in each learning cycle we train much fewer epochs with limited number of labeled data.
More importantly, compared to the annotation cost for an entire dataset (see Sec.~\ref{Sec:introducation} for an estimation on SHA),
ours is substantially reduced !

\medskip

\para{Acknowledgement: } This work was supported by the National Natural Science
Foundation of China (NSFC) under Grant No. 61828602 and 51475334; as well as National Key Research
and Development Program of Science and Technology of China under
Grant No. 2018YFB1305304, Shanghai Science and Technology Pilot
Project under Grant No. 19511132100.

\clearpage
\appendix


\section{Appendix: more results}
In this appendix, we offer more results on ShanghaiTech, UCF\_CC\_50, TRANCOS and DCC datasets.

\subsection{ShanghaiTech}
We first offer a variant of AL-AC inspired by~\cite{sinha2019iccv} and compare to our original AL-AC. \cite{sinha2019iccv} is a state of the art active learning method where a variational autoencoder (VAE) and an adversarial network are trained to play a min-max game discriminating between unlabeled and labeled data with domain label 0 and 1, respectively. Samples from those predicted as “unlabeled” with the lowest probabilities (near 0) are selected for active annotations. We find this min-max idea to be similar to the gradient reversal layer (GRL) in our proposed distribution alignment between labeled and unlabeled data. The GRL also assumes a domain label 0 for the unlabeled data and 1 for the labeled data. It multiplies the gradient by a negative constant (-1) during the network back propagation which enforces the feature distributions over the labeled and unlabeled data as indistinguishable as possible for the distribution classifier. We therefore select unlabeled samples with the lowest probabilities from our domain classifier. In this sense, the distribution alignment with latent MixUp is included in every learning cycle. We denote it by AL-AC-v as a variant of AL-AC and compare it to the full version of our AL-AC in the default setting (M = 40, m =10) on SHA and SHB in Table~\ref{Tab:Transfer}: Left. Our AL-AC still works clearly better than this variant.

Next, to test the generalization ability of our method, we offer the results under the default setting M = 40, m =10 by training on SHA and testing on SHB (SHA $\rightarrow$ SHB), and vice versa (SHB $\rightarrow$ SHA).
The MAE and MSE for our proposed AL-AC and baseline are reported in Table~\ref{Tab:Transfer}: Right. It can be seen that our AL-AL improves the baseline substantially in this transfer setting.

\begin{table}
	\setlength{\tabcolsep}{2pt}
	\centering
	\small
	\begin{tabular}{ccccc}
		\toprule
		M=40, m=10  & \multicolumn{2}{c}{SHA} & \multicolumn{2}{c}{SHB}\\
		\midrule
	 Method	 & MAE & MSE & MAE & MSE \\
	\midrule
		Baseline	&93.8	& 150.9	& 17.9	&27.3\\
		AL-AC-v  & 85.6	& 143.7 & 	14.8 &	23.7 \\
		AL-AC  & 80.4 &	138.8	&12.7&	20.4\\
		\bottomrule
	\end{tabular}
		\hspace{0.5cm}
			\begin{tabular}{ccccc}
		\toprule
		M=40, m=10  & \multicolumn{2}{c}{SHA $\rightarrow$ SHB} & \multicolumn{2}{c}{SHB $\rightarrow$ SHA}\\
		\midrule
	 Method	 & MAE & MSE & MAE & MSE \\
	\midrule
		Baseline	&39.2	&53.8	&167.4	&283.2 \\
		AL-AC  & 30.5	&45.0&	144.0	&238.5\\
 \bottomrule
	\end{tabular}
	\caption{\small Left: Comparison of AL-AC to its variant AL-AC-v inspired by~\cite{sinha2019iccv}. Right: Cross dataset performance of AL-AC. Experiments are on ShanghaiTech dataset with default setting $M= 40$ and $m =10$.   }
	\label{Tab:Transfer}	
		\vspace{-0.2cm}
\end{table}

\subsection{UCF\_CC\_50}
Our proposed AL-AC is mainly composed of two parts: 1) partition-based sample selection with weights (PSSW); and 2) distribution alignment with latent MixUp. We present detailed results of both components on the UCF\_CC\_50 dataset.

First, we compare our PSSW with random selection (RS) in Table~\ref{Tab:LabellingBudgetUCF}. We choose by default $M = 10$ and $m = 3$ (initial $m$ is 4). The mean MAE at the starting point ($M = 4, m = 4$) is 645.8 for both PSSW and RS. For PSSW, it reduces to 479.2 with $M = 7$, and $387.3$ with $M = 10$; in contrast, the MAE for RS is 505.8 and 444.7 for $M = 7$ and $M = 10$, respectively. PSSW produces clearly lower MAE than RS.  We also present the result of PSSW with $M = 20$ and $m = 10$: the MAE is also much lower than that of RS.

Next, we study the effect of the proposed distribution alignment with latent MixUp in Table~\ref{Tab:LabellingBudgetUCF}. Like in the paper, we add GRL (gradient reversal layer) and MX (latent MixUp) to PSSW and report the result. For $M = 10$, $m = 3$, by adding GRL + MX to PSSW, the mean MAE and MSE further reduce 35.9 and 58.8 points, respectively. We also present the result for $M = 20$, $m =10$, the contribution of GRL and MX is also significant (\eg 27.2  points decrease on MAE).
Notice PSSW + GRL + MX  is equivalent to AL-AC in Table 4.

\subsection{TRANCOS and DCC}
Previously, we present the result of AL-AC by labeling 10\% of images for TRANCOS and DCC datasets. In Table~\ref{tab:trancos-sup}, we present the result of labeling 20\% data for each; that is, $M = 160, m=20$ for TRANCOS and $M = 20, m =5$ for DCC. The mean MAE of AL-AC is 5.9 on TRANCOS with a decrease of 2.9
points from baseline; 3.8 on DCC with a decrease of 2.6
points from baseline. With 20\% labeled data, our AL-AC
performs quite close to the state of the art~\cite{marsden2018cvpr,li2018cvpr,shi2019iccv}, which utilize full annotations of the datasets.

\begin{table}[t]
	\setlength{\tabcolsep}{4pt}
	\centering
	\small
	\begin{tabular}{ccc}
		\toprule
		Dataset  & \multicolumn{2}{c}{UCF\_CC\_50}  \\
		\midrule
		Method & PSSW & RS \\
	\midrule
		M=4, m=4  &  645.8 \pms{36.5} & 645.8 \pms{36.5}   \\
		M=7, m=3&  479.2 \pms{32.1} & 505.8 \pms{35.3} \\
		M=10, m=3 & 387.3 \pms 22.5 & 444.7 \pms{25.9}   \\
		\midrule
		\midrule
		M=20, m=10  & 345.9 \pms 24.6 &  417.2\pms{29.8}\\
		\bottomrule
	\end{tabular}
		\hspace{0.5cm}
			\begin{tabular}{ccc}
		\toprule
		Dataset  & \multicolumn{2}{c}{UCF\_CC\_50}\\
		\midrule
		M=10, m=3 & MAE & MSE \\
	\midrule
			PSSW & 387.3   & 506.9   \\
		PSSW + GRL + MX     &   351.4  & 448.1  \\
	\midrule
	\midrule
		M=20, m=10 & MAE & MSE \\
	\midrule
		PSSW  &  345.9  & 498.3 \\
		PSSW+ GRL + MX     &   318.7  & 421.6\\
 \bottomrule
	\end{tabular}
	\caption{\small Ablation study of the proposed partition-based sample selection with weights (PSSW) and distribution alignment with latent MixUp (GRL + MX). Left: comparison of PSSW against random selection (RS), MAE is reported. Right: ablation of GRL + MX, MAE and MSE are reported. Experiments are on UCF\_CC\_50. }
	\label{Tab:LabellingBudgetUCF}	
		\vspace{-0.2cm}
\end{table}


\begin{table}
	\small
	\centering
		\begin{tabular}{c|cc|ccccc}
		\toprule
		\textsc{Method} & 	Baseline*  &		AL-AC* &  Lempitsky\cite{lempitsky2010nips} & Hydra-3s\cite{onoro2016eccv} & POCR~\cite{marsden2018cvpr} & 	 CSRNet~\cite{li2018cvpr}& 	CFF~\cite{shi2019iccv}  \\
		\midrule
		TRANCOS & 8.8 \pms 1.4 &  5.9 \pms 0.9 & 13.8 & 11.0 & 9.7 & 3.6 & 2.0  \\
		DCC    &  6.4 \pms 1.1 &  3.8 \pms 0.5 & - & - & 8.4 & -  & 3.2  \\
		\bottomrule
	\end{tabular}
	\caption{Comparison of AL-AC with state of the art on TRANCOS and DCC datasets. *Notice that the labeling budget of AL-AC on TRANCOS and DCC is different. We label 20\% of images over each set, such that $M=160, m=20$ for TRANCOS, and  $M=20, m=5$ for DCC. MAE is reported.}
	\label{tab:trancos-sup}
	\vspace{-0.2cm}
\end{table}
\begin{figure}[hbt!]
	\centering
	\includegraphics[width=1\columnwidth]{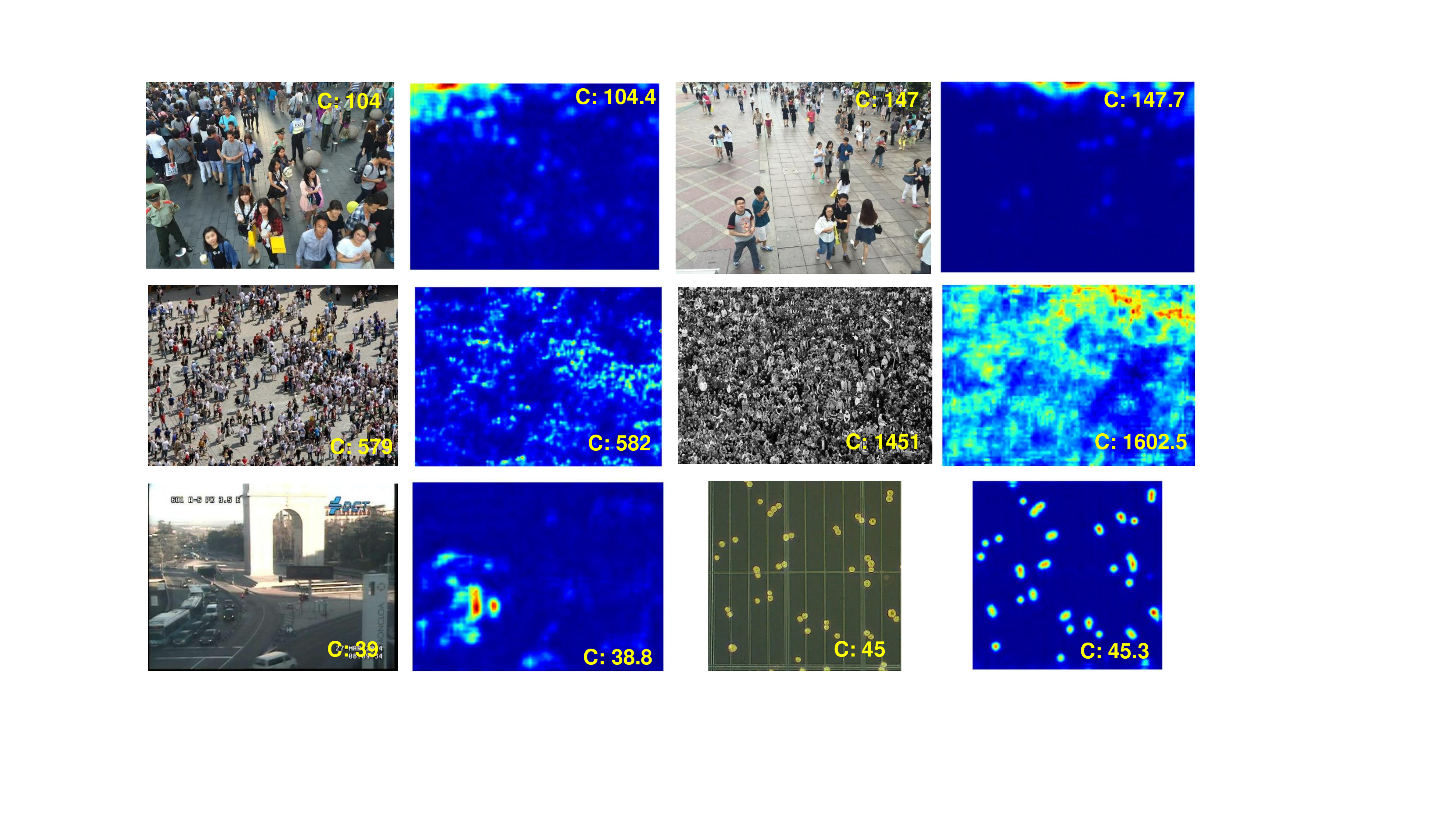}
	\caption{\small Examples of AL-AC. Ground truth counts are in the original images while predicted counts in the estimated density maps. }
	\label{Fig:Sup}
\end{figure}

\section{Appendix: more examples}
Several new examples of AL-AC are illustrated in Fig.~\ref{Fig:Sup} over different datsets (\eg ShanghaiTech, UCF\_CC\_50, TRANCOS, and DCC).

\clearpage
%
%

\bibliographystyle{splncs04}
\bibliography{aaai2019}

\begin{thebibliography}{10}
\providecommand{\url}[1]{\texttt{#1}}
\providecommand{\urlprefix}{URL }
\providecommand{\doi}[1]{https://doi.org/#1}

\bibitem{berthelot2019arxiv}
Berthelot, D., Carlini, N., Goodfellow, I., Papernot, N., Oliver, A., Raffel,
  C.: Mixmatch: A holistic approach to semi-supervised learning. arXiv preprint
  arXiv:1905.02249  (2019)

\bibitem{brostow2006cvpr}
Brostow, G.J., Cipolla, R.: Unsupervised bayesian detection of independent
  motion in crowds. In: CVPR (2006)

\bibitem{cao2018eccv}
Cao, X., Wang, Z., Zhao, Y., Su, F.: Scale aggregation network for accurate and
  efficient crowd counting. In: ECCV (2018)

\bibitem{change2013cvpr}
Change~Loy, C., Gong, S., Xiang, T.: From semi-supervised to transfer counting
  of crowds. In: CVPR (2013)

\bibitem{chen2012bmvc}
Chen, K., Loy, C.C., Gong, S., Xiang, T.: Feature mining for localised crowd
  counting. In: BMVC (2012)

\bibitem{dasgupta2005nips}
Dasgupta, S.: Analysis of a greedy active learning strategy. In: NIPS (2005)

\bibitem{ganin2014jmlr}
Ganin, Y., Lempitsky, V.: Unsupervised domain adaptation by backpropagation.
  In: JMLR (2015)

\bibitem{gonzalez2015cvpr}
Gonzalez-Garcia, A., Vezhnevets, A., Ferrari, V.: An active search strategy for
  efficient object class detection. In: CVPR (2015)

\bibitem{guerrero2015ibpra}
Guerrero-G{\'o}mez-Olmedo, R., Torre-Jim{\'e}nez, B., L{\'o}pez-Sastre, R.,
  Maldonado-Basc{\'o}n, S., Onoro-Rubio, D.: Extremely overlapping vehicle
  counting. In: Iberian Conference on Pattern Recognition and Image Analysis
  (2015)

\bibitem{hoffer2016arxiv}
Hoffer, E., Ailon, N.: Semi-supervised deep learning by metric embedding. arXiv
  preprint arXiv:1611.01449  (2016)

\bibitem{hossain2019wacv}
Hossain, M., Hosseinzadeh, M., Chanda, O., Wang, Y.: Crowd counting using
  scale-aware attention networks. In: WACV (2019)

\bibitem{hossainone2019bmvc}
Hossain, M.A., Kumar, M., Hosseinzadeh, M., Chanda, O., Wang, Y.: One-shot
  scene-specific crowd counting. In: BMVC (2019)

\bibitem{idrees2013cvpr}
Idrees, H., Saleemi, I., Seibert, C., Shah, M.: Multi-source multi-scale
  counting in extremely dense crowd images. In: CVPR (2013)

\bibitem{idrees2018eccv}
Idrees, H., Tayyab, M., Athrey, K., Zhang, D., Al-Maadeed, S., Rajpoot, N.,
  Shah, M.: Composition loss for counting, density map estimation and
  localization in dense crowds. In: ECCV (2018)

\bibitem{jenks1967data}
Jenks, G.F.: The data model concept in statistical mapping. International
  yearbook of cartography  \textbf{7},  186--190 (1967)

\bibitem{jiang2019cvpr}
Jiang, X., Xiao, Z., Zhang, B., Zhen, X., Cao, X., Doermann, D., Shao, L.:
  Crowd counting and density estimation by trellis encoder-decoder networks.
  In: CVPR (2019)

\bibitem{joshi2009cvpr}
Joshi, A.J., Porikli, F., Papanikolopoulos, N.: Multi-class active learning for
  image classification. In: CVPR (2009)

\bibitem{krizhevsky2012nips}
Krizhevsky, A., Sutskever, I., Hinton, G.E.: Imagenet classification with deep
  convolutional neural networks. In: NIPS (2012)

\bibitem{laine2017iclr}
Laine, S., Aila, T.: Temporal ensembling for semi-supervised learning. In: ICLR
  (2016)

\bibitem{laradji2018eccv}
Laradji, I.H., Rostamzadeh, N., Pinheiro, P.O., Vazquez, D., Schmidt, M.: Where
  are the blobs: Counting by localization with point supervision. In: ECCV
  (2018)

\bibitem{lee2013icmlw}
Lee, D.H.: Pseudo-label: The simple and efficient semi-supervised learning
  method for deep neural networks. In: ICMLW (2013)

\bibitem{lempitsky2010nips}
Lempitsky, V., Zisserman, A.: Learning to count objects in images. In: NIPS
  (2010)

\bibitem{li2018cvpr}
Li, Y., Zhang, X., Chen, D.: Csrnet: Dilated convolutional neural networks for
  understanding the highly congested scenes. In: CVPR (2018)

\bibitem{lian2019cvpr}
Lian, D., Li, J., Zheng, J., Luo, W., Gao, S.: Density map regression guided
  detection network for rgb-d crowd counting and localization. In: CVPR (2019)

\bibitem{liu2018cvpr}
Liu, J., Gao, C., Meng, D., G.~Hauptmann, A.: Decidenet: Counting varying
  density crowds through attention guided detection and density estimation. In:
  CVPR (2018)

\bibitem{liu2019cvpr}
Liu, W., Salzmann, M., Fua, P.: Context-aware crowd counting. In: CVPR (2019)

\bibitem{liu2019tpami}
Liu, X., Van De~Weijer, J., Bagdanov, A.D.: Exploiting unlabeled data in cnns
  by self-supervised learning to rank. IEEE transactions on pattern analysis
  and machine intelligence  (2019)

\bibitem{liu2018bcvpr}
Liu, X., Weijer, J., Bagdanov, A.D.: Leveraging unlabeled data for crowd
  counting by learning to rank. In: CVPR (2018)

\bibitem{liu2019cvprb}
Liu, Y., Shi, M., Zhao, Q., Wang, X.: Point in, box out: Beyond counting
  persons in crowds. In: CVPR (2019)

\bibitem{zhang2018wacv}
Lu, Z., Shi, M., Chen, Q.: Crowd counting via scale-adaptive convolutional
  neural network. In: WACV (2018)

\bibitem{ma2019iccv}
Ma, Z., Wei, X., Hong, X., Gong, Y.: Bayesian loss for crowd count estimation
  with point supervision. In: ICCV (2019)

\bibitem{marsden2018cvpr}
Marsden, M., McGuinness, K., Little, S., Keogh, C.E., O'Connor, N.E.: People,
  penguins and petri dishes: adapting object counting models to new visual
  domains and object types without forgetting. In: CVPR (2018)

\bibitem{olivier2006tnn}
Olivier, C., Bernhard, S., Alexander, Z.: Semi-supervised learning. In: IEEE
  Transactions on Neural Networks, vol.~20, pp. 542--542 (2006)

\bibitem{olmschenk2018wacv}
Olmschenk, G., Tang, H., Zhu, Z.: Crowd counting with minimal data using
  generative adversarial networks for multiple target regression. In: WACV
  (2018)

\bibitem{olmschenk2019cviu}
Olmschenk, G., Zhu, Z., Tang, H.: Generalizing semi-supervised generative
  adversarial networks to regression using feature contrasting. Computer Vision
  and Image Understanding  (2019)

\bibitem{onoro2016eccv}
Onoro-Rubio, D., L{\'o}pez-Sastre, R.J.: Towards perspective-free object
  counting with deep learning. In: ECCV (2016)

\bibitem{pham2015iccv}
Pham, V.Q., Kozakaya, T., Yamaguchi, O., Okada, R.: Count forest: Co-voting
  uncertain number of targets using random forest for crowd density estimation.
  In: ICCV (2015)

\bibitem{rabaud2006cvpr}
Rabaud, V., Belongie, S.: Counting crowded moving objects. In: CVPR (2006)

\bibitem{ranjan2018eccv}
Ranjan, V., Le, H., Hoai, M.: Iterative crowd counting. In: ECCV (2018)

\bibitem{rasmus2015nips}
Rasmus, A., Berglund, M., Honkala, M., Valpola, H., Raiko, T.: Semi-supervised
  learning with ladder networks. In: NIPS (2015)

\bibitem{sam2018aaai}
Sam, D.B., Babu, R.V.: Top-down feedback for crowd counting convolutional
  neural network. In: AAAI (2018)

\bibitem{sam2019aaai}
Sam, D.B., Sajjan, N.N., Maurya, H., Babu, R.V.: Almost unsupervised learning
  for dense crowd counting. In: AAAI (2019)

\bibitem{sam2017arxiv}
Sam, D.B., Surya, S., Babu, R.V.: Switching convolutional neural network for
  crowd counting. In: CVPR (2017)

\bibitem{sener2018iclr}
Sener, O., Savarese, S.: Active learning for convolutional neural networks: A
  core-set approach. In: ICLR (2018)

\bibitem{settles2009UWM}
Settles, B.: Active learning literature survey. Tech. rep., University of
  Wisconsin-Madison Department of Computer Sciences (2009)

\bibitem{shi2019cvpr}
Shi, M., Yang, Z., Xu, C., Chen, Q.: Revisiting perspective information for
  efficient crowd counting. In: CVPR (2019)

\bibitem{shi2019iccv}
Shi, Z., Mettes, P., Snoek, C.G.: Counting with focus for free. In: ICCV (2019)

\bibitem{sindagi2017iccv}
Sindagi, V.A., Patel, V.M.: Generating high-quality crowd density maps using
  contextual pyramid cnns. In: ICCV (2017)

\bibitem{sinha2019iccv}
Sinha, S., Ebrahimi, S., Darrell, T.: Variational adversarial active learning.
  In: ICCV (2019)

\bibitem{tan2011pr}
Tan, B., Zhang, J., Wang, L.: Semi-supervised elastic net for pedestrian
  counting. Pattern Recognition  \textbf{44}(10-11),  2297--2304 (2011)

\bibitem{verma2019icml}
Verma, V., Lamb, A., Beckham, C., Najafi, A., Mitliagkas, I., Courville, A.,
  Lopez-Paz, D., Bengio, Y.: Manifold mixup: Better representations by
  interpolating hidden states. In: ICML (2019)

\bibitem{verma2019arxiv}
Verma, V., Lamb, A., Kannala, J., Bengio, Y., Lopez-Paz, D.: Interpolation
  consistency training for semi-supervised learning. arXiv preprint
  arXiv:1903.03825  (2019)

\bibitem{viola2003ijcv}
Viola, P., Jones, M.J., Snow, D.: Detecting pedestrians using patterns of
  motion and appearance. IJCV  \textbf{63}(2),  153--161 (2003)

\bibitem{wang2016tcsvt}
Wang, K., Zhang, D., Li, Y., Zhang, R., Lin, L.: Cost-effective active learning
  for deep image classification. IEEE Transactions on Circuits and Systems for
  Video Technology  \textbf{27}(12),  2591--2600 (2016)

\bibitem{wang2019cvpr}
Wang, Q., Gao, J., Lin, W., Yuan, Y.: Learning from synthetic data for crowd
  counting in the wild. In: CVPR (2019)

\bibitem{weston2012nn}
Weston, J., Ratle, F., Mobahi, H., Collobert, R.: Deep learning via
  semi-supervised embedding. In: Neural Networks: Tricks of the Trade, pp.
  639--655. Springer (2012)

\bibitem{xiong2017iccv}
Xiong, F., Shi, X., Yeung, D.Y.: Spatiotemporal modeling for crowd counting in
  videos. In: ICCV (2017)

\bibitem{xu2019iccv}
Xu, C., Qiu, K., Fu, J., Bai, S., Xu, Y., Bai, X.: Learn to scale: Generating
  multipolar normalized density map for crowd counting. In: ICCV (2019)

\bibitem{yan2019iccv}
Yan, Z., Yuan, Y., Zuo, W., Tan, X., Wang, Y., Wen, S., Ding, E.:
  Perspective-guided convolution networks for crowd counting. In: ICCV (2019)

\bibitem{yang2015ijcv}
Yang, Y., Ma, Z., Nie, F., Chang, X., Hauptmann, A.G.: Multi-class active
  learning by uncertainty sampling with diversity maximization. International
  Journal of Computer Vision  \textbf{113}(2),  113--127 (2015)

\bibitem{yang2019arxiv}
Yang, Z., Shi, M., Avrithis, Y., Xu, C., Ferrari, V.: Training object detectors
  from few weakly-labeled and many unlabeled images. arXiv preprint
  arXiv:1912.00384  (2019)

\bibitem{zhang2015cvpr}
Zhang, C., Li, H., Wang, X., Yang, X.: Cross-scene crowd counting via deep
  convolutional neural networks. In: CVPR (2015)

\bibitem{zhang2018iclr}
Zhang, H., Cisse, M., Dauphin, Y.N., Lopez-Paz, D.: Mixup: Beyond empirical
  risk minimization. In: ICLR (2018)

\bibitem{zhang2016cvpr}
Zhang, Y., Zhou, D., Chen, S., Gao, S., Ma, Y.: Single-image crowd counting via
  multi-column convolutional neural network. In: CVPR (2016)

\bibitem{zhou2018tcsvt}
Zhou, Q., Zhang, J., Che, L., Shan, H., Wang, J.Z.: Crowd counting with limited
  labeling through submodular frame selection. IEEE Transactions on Intelligent
  Transportation Systems  \textbf{20}(5),  1728--1738 (2018)

\bibitem{zou2019bmvc}
Zou, Z., Shao, H., Qu, X., Wei, W., Zhou, P.: Enhanced 3d convolutional
  networks for crowd counting. In: BMVC (2019)

\end{thebibliography}
\end{document}